\documentclass[letterpaper, 10 pt, conference]{ieeeconf}
\makeatletter
\let\NAT@parse\undefined
\makeatother

\IEEEoverridecommandlockouts                              %

\usepackage[numbers,sort&compress]{natbib}

\usepackage{multirow}
\usepackage{multicol}
\usepackage[bookmarks=true]{hyperref}
\usepackage{xcolor}
\usepackage{tablefootnote} %
\usepackage{subcaption}
\usepackage{graphicx}
\usepackage{float}
\usepackage{amssymb}
\usepackage{amsmath}
\usepackage{tabularx}

\usepackage[capitalize]{cleveref}
\crefname{section}{Sec.}{Secs.}
\Crefname{section}{Section}{Sections}
\Crefname{table}{Table}{Tables}
\crefname{table}{Tab.}{Tabs.}

\usepackage{mathtools}

\def\eg{\emph{e.g.}}

\usepackage{tikz}

\usepackage{pifont}%
\newcommand{\cmark}{\ding{51}}%
\newcommand{\xmark}{\ding{55}}%

\usepackage{soul}
\let\labelindent\relax
\usepackage{enumitem}

\usepackage{amsmath,amsfonts,bm}

\def\eqref#1{equation~\ref{#1}}

\def\1{\bm{1}}

\def\rvc{{\mathbf{c}}}

\def\rvf{{\mathbf{f}}}

\def\rvq{{\mathbf{q}}}

\def\rmC{{\mathbf{C}}}

\def\rmF{{\mathbf{F}}}
\def\rmG{{\mathbf{G}}}

\def\rmI{{\mathbf{I}}}

\def\rmM{{\mathbf{M}}}

\def\rmP{{\mathbf{P}}}
\def\rmQ{{\mathbf{Q}}}

\def\rmS{{\mathbf{S}}}

\def\rmV{{\mathbf{V}}}

\DeclareMathAlphabet{\mathsfit}{\encodingdefault}{\sfdefault}{m}{sl}
\SetMathAlphabet{\mathsfit}{bold}{\encodingdefault}{\sfdefault}{bx}{n}

\begin{document}
\bstctlcite{BSTcontrol}

\title{\LARGE \bf
ODTFormer: Efficient Obstacle Detection and Tracking with Stereo Cameras Based on Transformer
}

\author{Tianye Ding$^{1*}$, Hongyu Li$^{2*}$, and Huaizu Jiang$^1$%
\thanks{This research is supported by the National Science Foundation under Award Number IIS-2310254.}
\thanks{$^{*}$Equal contribution.}%
\thanks{$^{1}$Northeastern University, Boston, MA, 02115. {\tt\small \{ding.tian, h.jiang\}@northeastern.edu}}%
\thanks{$^{2}$Brown University, Providence, RI, 02912. {\tt\small hli230@cs.brown.edu}}%
}

\maketitle

\newcommand{\hj}[1]{\textcolor{magenta}{Huaizu: #1}}
\newcommand{\lhy}[1]{\textcolor{red}{LHY: #1}}
\newcommand{\dty}[1]{\textcolor{blue}{Tianye: #1}}

\begin{abstract}
Obstacle detection and tracking represent a critical component in robot autonomous navigation. In this paper, we propose ODTFormer, a Transformer-based model that addresses both obstacle detection and tracking problems. For the detection task, our approach leverages deformable attention to construct a 3D cost volume, which is decoded progressively in the form of voxel occupancy grids. We further track the obstacles by matching the voxels between consecutive frames. The entire model can be optimized in an end-to-end manner. Through extensive experiments on DrivingStereo and KITTI benchmarks, our model achieves state-of-the-art performance in the obstacle detection task. We also report comparable accuracy to state-of-the-art obstacle tracking models while requiring only a fraction of their computation cost, typically ten-fold to twenty-fold less. 
Our code is available on \href{https://github.com/neu-vi/ODTFormer}{\texttt{https://github.com/neu-vi/ODTFormer}}.
\end{abstract}

\section{Introduction}

Obstacle detection and tracking represents a safety-critical challenge across various domains, including robot autonomous navigation~\cite{frey_locomotion_2022, chen_socially_2017, akmandor_deep_2022, zhao2e2, li2024stereonavnet} and self-driving vehicles~\cite{li_bevformer_2022, li_voxformer_2023, wei_surroundocc_2023, huang_tri-perspective_2023}.
For instance, a service robot needs to detect people and pillars surrounding it, track their motions (if any), or even predict their future trajectories to avoid collision.
Accurate obstacle detection and tracking are crucial components of autonomous navigation systems, particularly in state-based frameworks, to ensure collision-free navigation~\cite{eppenberger_leveraging_2020, xu_real-time_2023, lu_perception_2022, wang_autonomous_2021}.
Recent research efforts focus on using low-cost visual sensors for obstacle perception to improve affordability~\cite{eppenberger_leveraging_2020, li2023stereovoxelnet, kareer_vinl_2023, zhou_raptor_2020, loquercio_learning_2021} compared with costly ones (\eg, LiDAR).
In this paper, we concentrate on a specific line of research employing stereo cameras, which offer higher 3D perception accuracy, extended sensing range, and enhanced agility for robots compared to monocular-based systems~\cite{falanga_how_2019, wang_pseudo-lidar_2019}.

\begin{figure}[!t]
    \centering
    \includegraphics[width=\columnwidth]{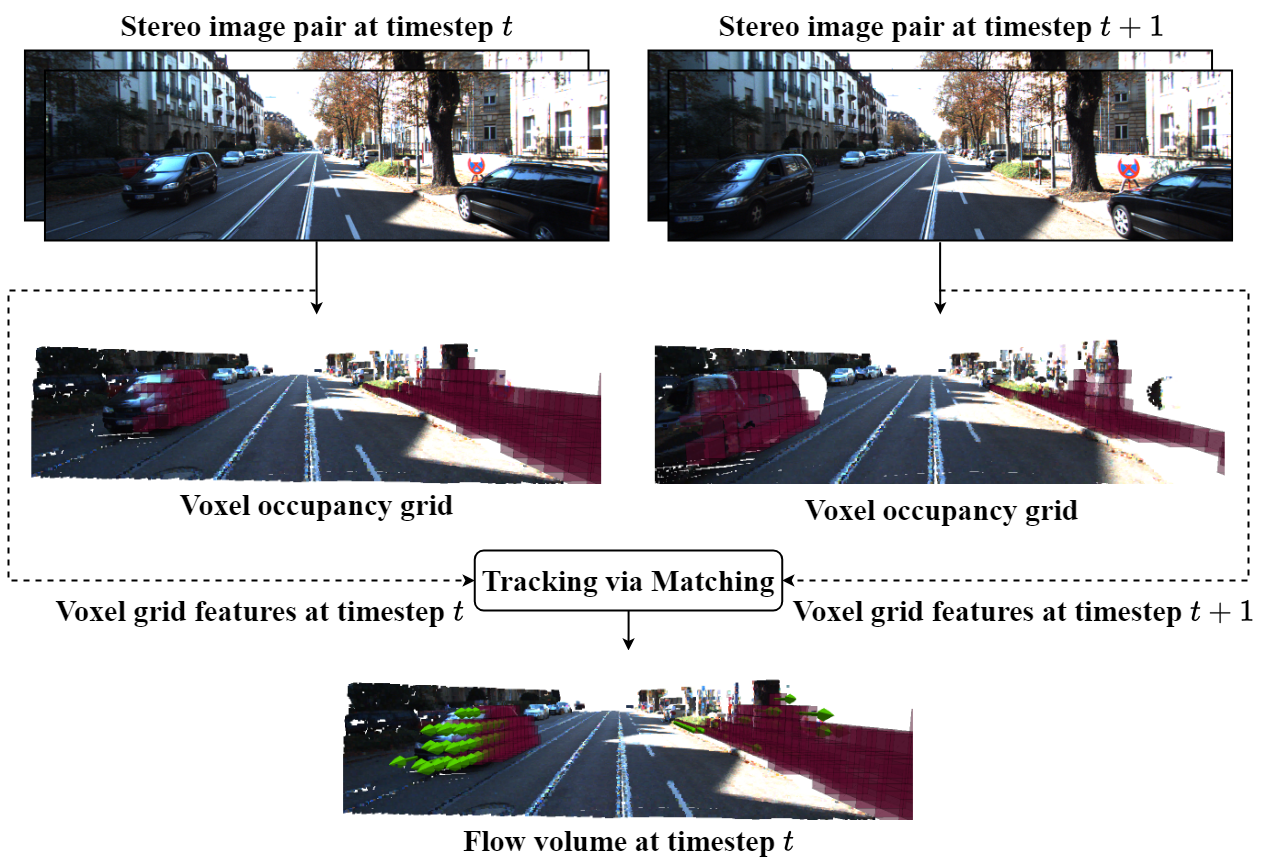}
    \caption{\textbf{We propose ODTFormer for joint obstacle detection and tracking using stereo cameras.}
    We first detect obstacles in the form of occupancy grids at each time step and match them across two consecutive frames for tracking. 
    We can see here that our model can successfully detect all obstacles and accurately track them.
    The obstacle detection results are shown as red cubes, and the tracking results are marked as green arrows.
Longer arrows indicate large motion magnitude.
    }
    \label{fig:teaser}
\end{figure}

Previous stereo-based obstacle detection approaches have mostly relied on depth estimation modules~\cite{eppenberger_leveraging_2020, li2023stereovoxelnet, loquercio_learning_2021}.
These approaches involve the estimation of a depth map followed by its transformation into a point cloud or a voxel grid. 
However, this two-stage approach often necessitates tradeoffs between speed and accuracy.
\citet{li2023stereovoxelnet} addressed this tradeoff by introducing StereoVoxelNet, which directly estimates voxel grids from stereo images by constructing a pixel-wise cost volume and employing a 2D-3D encoder-decoder network structure for efficient and accurate voxel grid reconstruction. 
However, a notable limitation of StereoVoxelNet is its \textit{implicit} incorporation of camera parameters within the neural network parameters.
As a result, the model can only work with specific camera parameters and image resolution, limiting its applicability across datasets with varying resolutions and camera parameters.

Moreover, the assumption of static obstacles presents challenges in dynamic environments, such as social navigation, where obstacles may be randomly moving pedestrians.
Numerous studies have addressed the visual dynamic obstacle tracking problem, employing traditional methods such as the Kalman filter to estimate the velocity~\cite{xu_real-time_2023, wang_autonomous_2021}, associating point cloud centroids~\cite{wang_autonomous_2021, eppenberger_leveraging_2020} or bounding boxes~\cite{lu_perception_2022} from consecutive frames using distance thresholds.
However, these techniques often require meticulous parameter tuning, leading to a lack of robustness.
Stereo-based scene flow estimation
addresses simultaneous 3D structure and motion estimation~\cite{vedula_three-dimensional_2005}, which could be adapted for obstacle tracking.
Nonetheless, these methods estimate a dense motion vector map and depth map for each pixel, resulting in high computational costs and runtime~\cite{jiang_sense_2019, teed_raft-3d_2021}.

In this paper, we present ODTFormer, a Transformer-based~\cite{vaswani_attention_2017} model, to address the aforementioned \underline{o}bstacle \underline{d}etection and \underline{t}racking challenges.
Unlike existing stereo-based models, which construct a pixel-wise cost volume for disparity matching~\cite{shamsafar_mobilestereonet_2022, shen_cfnet_2021, liu_local_2022, xu_attention_2022, li2023stereovoxelnet}, we propose to use deformable cross-attention~\cite{zhu2020deformable} from 3D voxel queries to 2D stereo image features to compute the matching cost.
Compared with the pixel-wise cost volume used in~\cite{li2023stereovoxelnet}, we directly construct it in the 3D space, conforming better to the scene geometry.
More importantly, 
our approach disentangles dataset specifics from the model design and thus shows better generalization than~\cite{li2023stereovoxelnet}.
The cost volume is then processed by a U-Net decoder to produce occupancy voxel grids progressively for efficiency purposes.

To account for environmental dynamics, we introduce a novel obstacle tracker to capture the motions of the scene by matching similar voxels between two consecutive frames.
We integrate physical constraints into the voxel tracking by setting a volumetric bound for each voxel when searching for its corresponding voxel in the next frame, which improves both accuracy and efficiency.
On the one hand, compared with traditional approaches, our model has the learning capacity that leverages the deep voxel feature representations to track obstacles accurately.
Moreover, both detection and tracking modules can be jointly optimized, leading to high accuracy.
On the other hand, compared with scene flow methods, our approach tracks the sparse voxels instead of densely over pixels, leading to lightweight computation with higher efficiency.
Our entire model can detect and track obstacles efficiently, resulting in as much as 20 times fewer computations compared to the current state-of-the-art in scene flow estimation~\cite{teed_raft-3d_2021}.

To validate the effectiveness of our approach, we conduct comprehensive evaluations on DrivingStereo~\cite{yang2019drivingstereo} and KITTI~\cite{menze2015object} datasets.
In the obstacle detection task, we demonstrate a significant performance improvement compared to prior works~\cite{shamsafar_mobilestereonet_2022, shen_cfnet_2021, liu_local_2022, xu_attention_2022, li2023stereovoxelnet}.
Additionally, our approach exhibits greater generalization across various camera parameters and resolutions than~\cite{li2023stereovoxelnet}.
In the obstacle tracking task, we compare scene flow-based approaches and show that our approach can achieve comparable results to state-of-the-art methods while needing much lighter computation cost.
Detailed ablation studies are conducted to validate different design choices.
In summary, our contributions are:
\begin{itemize}[noitemsep,topsep=2pt,leftmargin=*]%
    \item A novel 3D cost volume construction method based on the deformable cross-attention~\cite{zhu2020deformable}, which better conforms to the scene geometry and generalizes well to different camera parameters and image resolutions.
    \item A novel obstacle tracking method by matching the voxels across two consecutive frames, which can be jointly optimized with the detection module in an end-to-end manner.
    \item Experimental results show that our approach achieves better or comparable detection and tracking accuracy to state-of-the-art methods while being efficient and running fast at the same time (20fps on the KITTI resolution of 370$\times$1224 on an RTX A5000 GPU without careful postprocessing, \eg, quantization, pruning).
\end{itemize}

\section{Related Works}
In this section, we introduce two related areas to our work.
We first introduce the general background of stereo vision, including depth and scene flow estimation.
We then survey the obstacle perception works from the perspectives of detection and tracking.

\begin{figure*}[t!]
\vspace*{0.15cm}
\centering
\includegraphics[width=\linewidth]{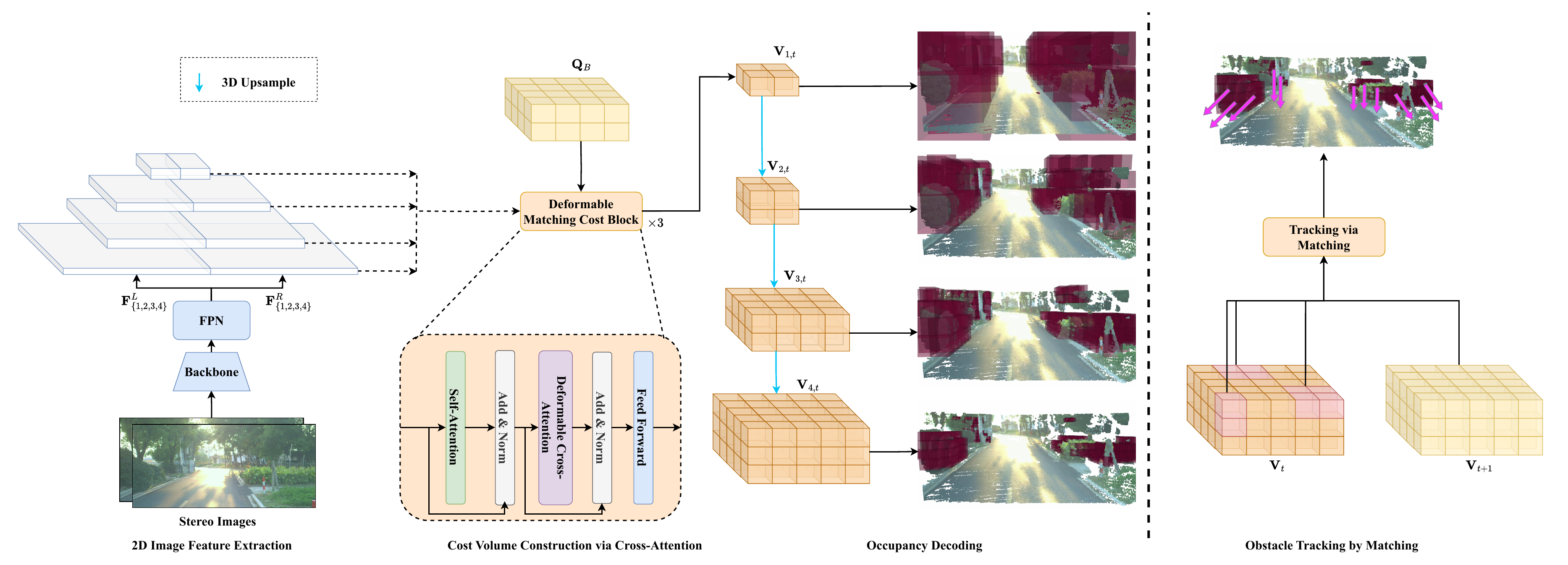}
\caption{\textbf{Illustration of the overall architecture design. }
\textbf{Left:} For obstacle detection, we first extract multi-scale 2D feature maps~\cite{tan_efficientnet_2019, lin2017feature} for each of the stereo images.
We then encode the voxels in the ROI to cross-attend to the image features to compute the matching cost through our novel cost volume construction method. 
Such a cost volume is directly constructed in the 3D space, which conforms better to the scene geometry, disentangles dataset specifics from model design, and thus generalizes well.
It is then progressively decoded into occupancy voxel grids.
\textbf{Right:} For obstacle tracking, we cast it as a matching problem by finding the correspondences of voxels across two consecutive frames, where we incorporate physical constraints to improve both the accuracy and efficiency.
Both the detection and tracking modules can be jointly optimized in an end-to-end manner and run efficiently.
}
\label{fig:op}
\end{figure*}

\subsection{Stereo Vision}
Stereo vision is a classical computer vision problem that aims to extract 3D information from a pair of RGB cameras.
Typically, the transformation between left and right cameras is considered rigid and known.
We briefly introduce two relevant applications of stereo vision: depth estimation and scene flow estimation.

Depth estimation involves deriving a depth map from one image of the stereo pair, typically the left one, associating each pixel with a depth value.
This is performed by disparity matching between the stereo pair~\cite{hirschmuller_accurate_2005}.
Recent approaches employ end-to-end deep neural networks and cost volume techniques for matching~\cite{shamsafar_mobilestereonet_2022, shen_cfnet_2021, liu_local_2022, xu_attention_2022, li2023stereovoxelnet}.

Scene flow estimation aims to understand the dynamic scenes by extracting the 3D information (disparity) and the motion simultaneously from sequences of stereo images~\cite{saxena2019pwoc, jiang_sense_2019,teed_raft-3d_2021}.
Unlike the traditional optical flow problem, scene flow estimation provides a denser correspondence between consecutive frames.
Stereo-based methods typically separate the 3D flow problem into depth and flow estimation.
Some works propose simultaneous estimation of 2D optical flow and disparity maps~\cite{jiang_sense_2019}, while others~\cite{teed_raft-3d_2021} leverage off-the-shelf depth estimators.

Despite their impressive performance, these methods often come with significant computational costs and are, therefore, unsuitable for real-time robotic systems due to the need for dense 2D estimation.
Our approach leverages the sparsity of these tasks and voxel grids, a common representation in robotics, significantly reducing computational costs while maintaining accuracy comparable to depth and scene flow methods.

\subsection{Obstacle Perception}
Obstacle perception is a fundamental aspect of robotics, playing a crucial role in enabling autonomous robots to navigate among environments safely and effectively. 
In this paper, we consider two aspects of obstacle perception: obstacle detection and obstacle tracking.

In the obstacle detection task, we detect obstacles in the form of a voxel occupancy grid.
Conventionally, stereo-based obstacle detection relies on filtering and voxelizing point clouds obtained by depth estimation module~\cite{eppenberger_leveraging_2020, shepel2021occupancy} but suffers from high computation complexity.
StereoVoxelNet~\cite{li2023stereovoxelnet} directly infers occupancy grids from stereo pairs but struggles with varying resolutions and camera parameters. 

Obstacle tracking aims at estimating the motion of obstacles across time.
While most of the previous works are based on LiDAR point cloud~\cite{wu2022casa, wu2023virtual} or camera-LiDAR fusion~\cite{wang2023you}, visual end-to-end methods, especially stereo-based and occupancy-based ones, remain under-explored.
Existing visual approaches rely on object detectors or Kalman filters~\cite{eppenberger_leveraging_2020, lu_perception_2022, xu_real-time_2023}, limiting generalizability and requiring meticulous parameter tuning.
We introduce voxel flow, which is the 3D grid-to-grid voxel-level motion of occupied voxels while achieving near real-time obstacle tracking.

In parallel to our work, there exist seminal occupancy detection approaches tailored for autonomous driving~\cite{wei_surroundocc_2023, li_voxformer_2023}.
These approaches commonly operate under the assumption of having access to images from the surrounding views. 
In contrast, our approach utilizes only a pair of stereo images and emphasizes computational efficiency.

\section{Proposed Method}
In this paper, we present a novel approach to tackle both obstacle detection and tracking problems.
For the obstacle detection task, our model estimates a voxel occupancy grid in front of the stereo cameras, defining this area as our region of interest (ROI).
Subsequently, our model efficiently tracks the motion of each detected obstacle by matching voxels across two consecutive frames.

\subsection{Overview}
\label{sec:overview}
\textbf{Obstacle detection.} Following StereoVoxelNet~\cite{li2023stereovoxelnet}, we formulate obstacle detection as occupancy detection based on stereo images.
As shown in Fig.~\ref{fig:op} (a), our detection model consists of three modules: i) 2D image feature extraction, ii) cost volume construction via cross-attention, and iii) occupancy decoding.

\subsubsection{2D image feature extraction} 
Given the rectified stereo images $\rmI^L$ and $\rmI^R$, we employ a lightweight EfficientNet-B0~\cite{tan_efficientnet_2019} backbone enhanced by the FPN~\cite{lin2017feature} to extract multi-scale image features, denoted as $\rmF_i^L$ and $\rmF_i^R$, respectively.
Here the subscript $i\in{\{1, 2, 3, 4\}}$ indicates the level (scale) of the feature maps unless explicitly mentioned otherwise.
With the input image resolution denoted as $H \times W$, the multi-scale feature maps reduce the resolutions to $\{\frac{1}{4}, \frac{1}{8}, \frac{1}{16}, \frac{1}{32}\}$, with the feature channel dimension of $D$ for all the levels.

\subsubsection{Cost volume construction via cross-attention} 
A crucial difference of our proposed approach from~\cite{li2023stereovoxelnet} is our novel cost volume construction module design, where we employ cross-attention from 3D voxel queries to 2D stereo image features.
Compared with the pixel-wise cost volume used in~\cite{li2023stereovoxelnet}, we directly construct it in the 3D space, conforming better to the scene geometry.
More importantly, 
our approach disentangles dataset specifics from the model design and thus shows better generalization than~\cite{li2023stereovoxelnet}.
We will introduce this module in detail in \cref{subsec:3d_2d_aggregation}.

\subsubsection{Occupancy decoding} 
\label{sec: occ_decoding}
Similar to~\cite{li2023stereovoxelnet}, as the final step in the obstacle detection module, we estimate the voxel occupancy in a coarse-to-fine-grained manner.
We utilize a 3D U-Net decoder to upsample the cost volume gradually.
Specifically, in the time step $t$,
our model outputs four-level coarse-to-fine voxel occupancy volumes $\rmV_{i,t}$ with the constant ROI size of $s_x \times s_y \times s_z$ meters,
\begin{equation}
\rmV_{i,t} = \{0,1\}^{2^{i - 1} (n_x \times n_y \times n_z)}, i \in \{1, 2, 3, 4\},
\end{equation}
where $i$ denotes the level number. $n_x$, $n_y$, and $n_z$ denote the number of voxels in the $x$, $y$ and $z$ axes, respectively.
Each voxel is classified as either empty (denoted by $0$) or occupied (denoted by $1$) and has a side length of $l_{i}$ which is decreased by half when the level number increases to keep grid metric size $s_{\{x,y,z\}}$ constant, $l_{i+1} = \frac{1}{2}l_i$.

\textbf{Obstacle tracking by matching.} After getting obstacle detection results in the form of occupied voxels, we'd like to track each of them for a better understanding of scene dynamics.
To this end, we find correspondences of occupied voxels across two consecutive time steps via \emph{global matching}, as shown in Fig.~\ref{fig:op} (the right part). 
Such tracking information provides critical cues of dynamic objects for visual navigation.
We introduce this part in detail in \cref{subsec:tracking}.

Finally, we describe how the entire model is trained in \cref{subsec:model_training}.

\subsection{Cost Volume Construction via Cross-Attention} 
\label{subsec:3d_2d_aggregation}
Cost volume plays a critical role for stereo-based obstacle detection methods~\cite{shamsafar_mobilestereonet_2022, shen_cfnet_2021, liu_local_2022, xu_attention_2022, li2023stereovoxelnet}.
Conventional cost volume construction involves computing \emph{pixel-wise} matching costs across a predefined range of disparity/depth levels~\cite{li2023stereovoxelnet}.
However, this approach encounters significant drawbacks. 
First of all, the dimensions of the cost volume do not directly correlate with the resolution of the voxel grid.
To solve this issue, the cost volume is compressed into a single latent vector and subsequently resized into the desired ROI voxel volume in~\cite{li2023stereovoxelnet}.
This process inevitably leads to the loss of geometry information, which is critical for accurate obstacle detection.
Furthermore, the hallucination of the latent vector to a 3D voxel volume assumes a fixed ROI dimension, prohibiting its generalization to a different setting\footnote{It is in analogy to the existence of the fully connected layers in AlexNet. As a result, the model only accepts an input image of a fixed dimension.}.
In addition, the camera intrinsics and extrinsics are directly embedded into the construction of the cost volume.
Such entanglement of the dataset specifics and model design restricts its applicability across different domains and setups, making generalization challenging, especially in scenarios with varying camera parameters and image resolutions.

Instead of constructing a cost volume at the pixel level, we propose a novel approach to build it in the 3D space directly via cross-attention from voxel queries to image features.
In a nutshell, we first encode each voxel in the ROI, which will then attend to image features via deformable attention~\cite{zhu2020deformable}.
Compared with the conventional cost volume~\cite{li2023stereovoxelnet}, our approach better conforms to the scene geometry and disentangles dataset specifics and model design, allowing better generalization.
We explain different components in detail as follows.
For brevity, we omit the time step index $t$ in this section.

\textbf{Encoding of voxels in the ROI.}
We partition the ROI into a set of voxels $\rmQ_B$ with a dimension of $n_x \times n_y \times n_z$ to balance the computation burden and spatial resolution of the cost volume.
Each voxel in $\rmQ_B$ is encoded as the Fourier positional encoding~\cite{vaswani_attention_2017} of its normalized centroid coordinates (between 0 and 1), which is fed into a multi-layer perceptron (MLP) to get the encoding $\rvq$.
It is further enhanced by the pixel-aligned features $\rvf$,  $\rvq=\rvq+\rvf$ as in~\cite{xie_pixel-aligned_2023}. $\rvf$ is defined as 
\begin{equation}
    \rvf = \frac{1}{2}(\rmF_4^L(\mathcal{P}(p, \theta^{L})) + \rmF_4^R(\mathcal{P}(p,\theta^{R}))),
\end{equation}
where $p$ is the voxel's centroid. 
$\mathcal{P}(p, \theta^{L})$ and $\mathcal{P}(p, \theta^{R})$ denote the projections of the voxel centroid onto the left and right images, respectively, according to their projection matrices $\theta^L$ and $\theta^R$. 
Bilinear interpolation is used to sample features from the image feature maps $\rmF_4^L$ and $\rmF_4^R$.

\textbf{Deformable cross-attention for matching cost.} 
We leverage the deformable matching cost (DMC) block to update voxel queries with calculated matching costs for each query.
Our DMC block is built on top of the transformer decoder block~\cite{vaswani_attention_2017} with deformable attention~\cite{zhu2020deformable}.

Specifically, for the voxel centered at $p$, we first sample points on multi-scale feature maps $(\rmF_i^L, \rmF_i^R)$ (similar to the pixel-aligned features $\rvf$ but on multi-scale features instead of just the last level) around the projected points of the voxel centroid $p$ with the offset $\delta$, respectively.
$\delta$ is generated by a learnable 3D offset sampler $g$~\cite{zhu2020deformable} based on the voxel's encoding $\rvq$, $\delta=g(\rvq)$, where $g$ is an MLP plus a sigmoid output layer (scaled by the 3D voxel size of the coarsest output, see experiment settings in Sec.~\ref{sec: ob-detection}).
We then directly concatenate the sampled features along the channel dimension and employ a scale-specific MLP to compute the cross-view matching cost $\rvc_i$:
\begin{equation}
\begin{aligned}
    \rvc_i(p, \rvq, \rmF_i) = \texttt{MLP}_i(&\rmF_i^L(\mathcal{P}(p + \delta, \theta^{L})) \oplus \\ 
    & \rmF_i^R(\mathcal{P}(p + \delta, \theta^{R}))),
    \label{eq:matching_cost}
\end{aligned}
\end{equation}
where $\oplus$ denotes the feature concatenation.
$\texttt{MLP}_i$ has two linear layers with batch normalization and ReLU activation in between. 
Since we project the same 3D sample point onto multi-scale feature maps, the resultant multi-scale matching costs for this 3D point can be aggregated by concatenating along the channel dimension and employing an MLP to compute the voxel matching cost $\rvc$:
\begin{equation}
\begin{aligned}
    \rvc(p, \rvq, \rmF_{\{1, 2, 3, 4\}}) = 
    \texttt{MLP}(& \rvc_1(p, \rvq, \rmF_1) \oplus \rvc_2(p, \rvq, \rmF_2) \oplus \\
    & \rvc_3(p, \rvq, \rmF_3) \oplus \rvc_4(p, \rvq, \rmF_4)),
    \label{eq:matching_cost_distill}
\end{aligned}
\end{equation}
where $\texttt{MLP}$ has the same layer configuration as $\texttt{MLP}_i$.
Instead of using a single offset, multiple ones are usually generated to enhance the computing of the matching cost, and different matching costs for different samples are aggregated as follows.
\begin{equation}
    \rmC_B(p, \rvq, \rmF) = \sum_{s = 1}^{N_s}\mathbf{A}_{s}\mathbf{W}_s \rvc(p, \delta_s, \rmF_{\{1, 2, 3, 4\}}),
\end{equation}
where $N_s$ is the total number of sampling points. $\delta_s$ denotes the $s$-th learned offset. 
$\mathbf{A}_{s} \in [0, 1]$ is the learnable weights for the cost generation and $\mathbf{W}_s \in \mathbb{R}^{D \times D}$ are learnable parameters.
We can get a cost volume $\rmC_B\in\mathbb{R}^{n_x\times n_y\times n_z\times D}$ by computing the matching cost for all voxels in $\rmQ_B$.

Deformable cross-attention has been used in previous works~\cite{li_bevformer_2022, li_voxformer_2023, wei_surroundocc_2023, huang_tri-perspective_2023} to aggregate multi-view and multi-scale features.
But instead of simply averaging them as in existing work, we leverage the geometric constraints between two stereo images and use the deformable attention to construct a cost volume in 3D, leading to better accuracy, as evidenced by our ablation studies.

The cost volume $\rmC$ will be processed to output occupancy voxel grid $\rmV_{i,t}$ progressively using a U-Net decoder~\cite{li2023stereovoxelnet}.

\subsection{Voxel Tracking by Matching}
\label{subsec:tracking}
Robots usually navigate in a dynamic environment by independently moving objects around (\eg, pedestrians).
Enabling a robot to track an object is essential to ensure safe navigation (\eg, avoiding bumping into a person).

Given the estimated occupancy grid $\rmV_{t}$ and $\rmV_{t+1}$ at the time step $t$ and $t+1$, respectively\footnote{They are the output of the last layer in the U-Net decoder, $\rmV_{4,t}$ and $\rmV_{4,t+1}$. For brevity, we simply use $\rmV_t$ and $\rmV_{t+1}$ in this section when the context is clear.}, our goal is to find the 3D motion vector of each voxel in $\rmV_{t}$ by finding its correspondence in $\rmV_{t+1}$.
Apparently, we do not need to worry about unoccupied voxels.
For the occupied ones, we compute their cosine similarities $\rmS_t\in[-1, 1]^{N_t\times N_{t+1}}$ by comparing their feature representations in the U-Net decoder right before the output layer. 
$N_t$ and $N_{t+1}$ denote the number of occupied voxels in $\rmV_t$ and $\rmV_{t+1}$, respectively.
Optionally, we can match all voxels in $\rmV_{t+1}$ regardless of their occupancy status. We will ablate this design choice in the experiment section.
The matching distribution of the occupied voxels can then be computed as
\begin{align}
    \rmP_t = \text{softmax}(\tau\rmS_t),
\end{align}
where $\tau$ is a learnable logit scale. The softmax is applied in a row-wise manner.
Motion vectors $\rmM_t$ of the occupied voxels can be computed as
\begin{align}
    \rmM_t = \rmP_t \rmG_{t+1} - \rmG_{t},
\end{align}
where $\rmG_t\in\mathbb{Z}^{N_t\times 3}$ denotes the 3D coordinates of each voxel's centroid in $\rmV_t$. 

In practice, the motions of the occupied voxels are bounded by physical constraints. 
Instead of naively comparing all possible pairs of occupied voxels, we set a volumetric bound 
surrounding each occupied voxel in $\rmV_t$ and only measure the similarities for voxels contained in this boundary when computing $\rmS_t$.
For the voxels outside of the boundary, the similarity score is set to be $-\infty$.
Specifically, assuming that a stereo image sequence is captured at least $f$ frames per second (FPS) and the obstacles have a smaller relative velocity of than $v$ meters per second to us, the maximum displacement of any trackable voxel across two frames is $d = \frac{v}{f}$.
With the voxel at the center of the volumetric region, the volume bound's dimensionality can be calculated as $(2\lceil \frac{d}{l_4} \rceil + 1) \times 3 \times (2\lceil \frac{d}{l_4} \rceil + 1)$ for $x$, $y$ and $z$ axes, where $l_4$ is the voxel side length in the last level of the U-Net decoder. The maximum displacements along the $y$-axis (pointing downward the ground plane) in both directions have been set to 1 to ignore drastic vertical movements within the driving scenario.
As we will see in the experiment section, it not only eases the computation burden but also leads to better obstacle tracking accuracy.

\subsection{Model Training} 
\label{subsec:model_training}
We optimize the obstacle detection model using the weighted sum of the Intersection of Union (IoU) loss $\mathcal{L}_D$ over four U-Net decoder levels:
\begin{align}
    \mathcal{L}_D = \sum_{j = 1}^{4} w_j(1 - \texttt{IoU}(\rmV_{j,t}, \bar{\rmV}_{j,t})),
\end{align}
where $\bar{\rmV}_{j,t}$ denotes the ground-truth voxel grid at level $j$.
$w_j$ is the weight assigned to the U-Net decoder level, and we set $w_1$ to $w_4$ as $[0.30, 0.27, 0.23, 0.20]$ empirically.

The tracker is trained using the endpoint error (EPE) loss, which is the average of element-wise L2 distances between all estimated and ground-truth 3D motion vectors.
We note, however, that optimizing the loss for detected occupied voxels leads to degenerated performance because the detection results may be wrong.
Instead, we define the loss for \emph{all} voxels regardless of their detection results.
The loss is defined as 
\begin{align}
    \mathcal{L}_T = \frac{1}{N'_{t}}\sum^{N'_{t}}_{k = 1}\big|\big|\rmM'_t[k] - \rmM_t[k]\big|\big|_2,
\end{align}
where $\rmM'$ contains the motion vectors for all the voxels.
For the $k$-the voxel, if it is detected as occupied, $\rmM'[k] = \rmM[k]$.
Otherwise, $\rmM'[k]=\mathbf{0}$.
$\bar{\rmM}'$ denotes the ground-truth motion vectors.

The ODTFormer is trained progressively.
We first train the detection module to obtain reliable detection results. We then jointly train the detection and tracking modules together, where the feature representations of the voxels are refined to encode the spatio-temporal cues effectively.

\section{Experiments}
In this section, we evaluate the performance of ODTFormer against various baselines on both obstacle detection (\cref{sec: ob-detection}) and obstacle tracking (\cref{sec: ob-tracking}) tasks.
In the end, we perform ablation studies on our design choices for both tasks (\cref{sec: ablation}).

\begin{figure*}[!htb]
\vspace*{0.15cm}
\centering
\begin{minipage}[t]{0.19\textwidth}
\centering
\includegraphics[width=\linewidth]{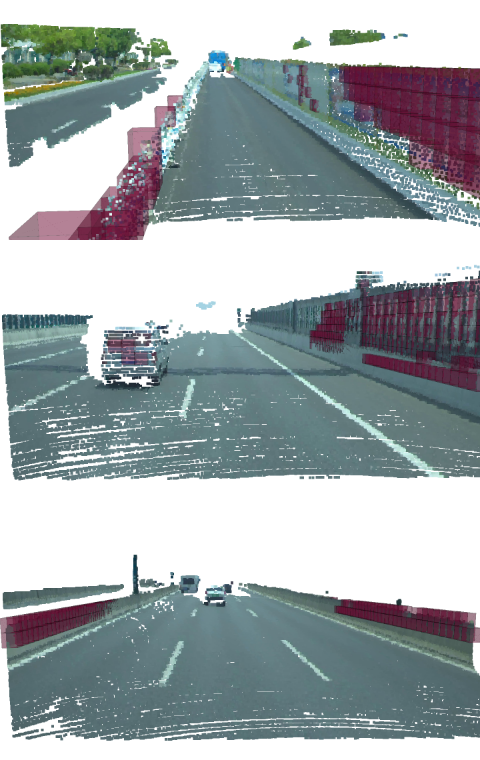}
Ground Truth
\end{minipage}
\begin{minipage}[t]{0.19\textwidth}
\centering
\includegraphics[width=\linewidth]{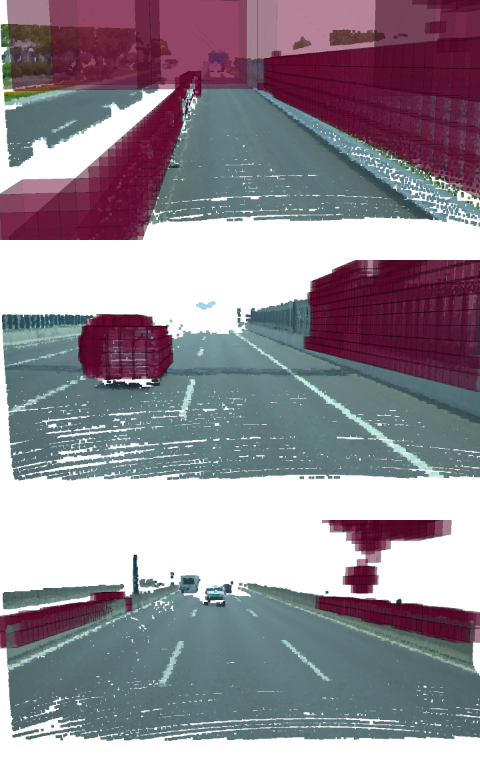}
3D-MSNet~\cite{shamsafar_mobilestereonet_2022}
\end{minipage}
\begin{minipage}[t]{0.19\textwidth}
\centering
\includegraphics[width=\linewidth]{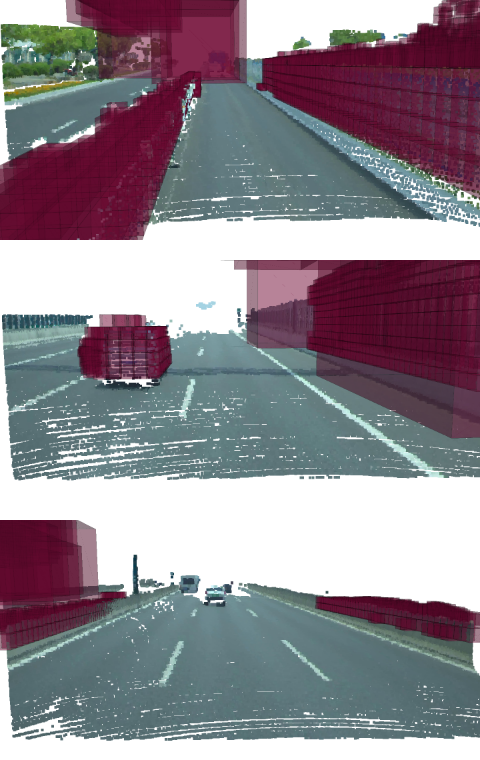}
Lac-GwcNet~\cite{liu_local_2022}
\end{minipage}
\begin{minipage}[t]{0.19\textwidth}
\centering
\includegraphics[width=\linewidth]{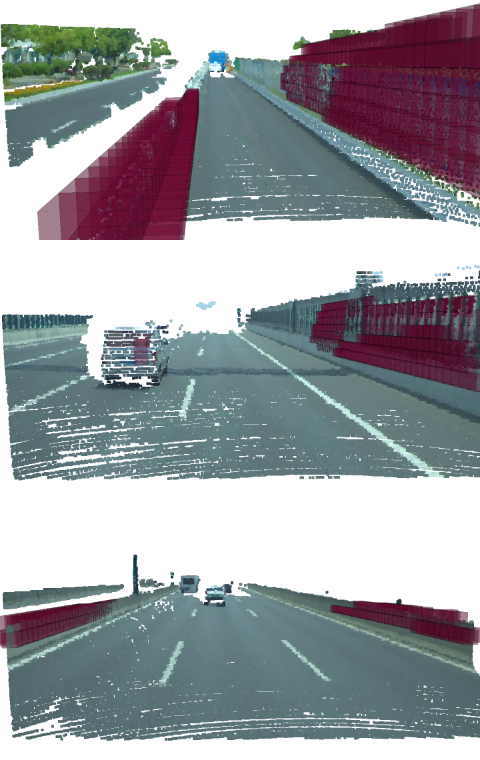}
StereoVoxelNet~\cite{li2023stereovoxelnet}
\end{minipage}
\begin{minipage}[t]{0.19\textwidth}
\centering
\includegraphics[width=\linewidth]{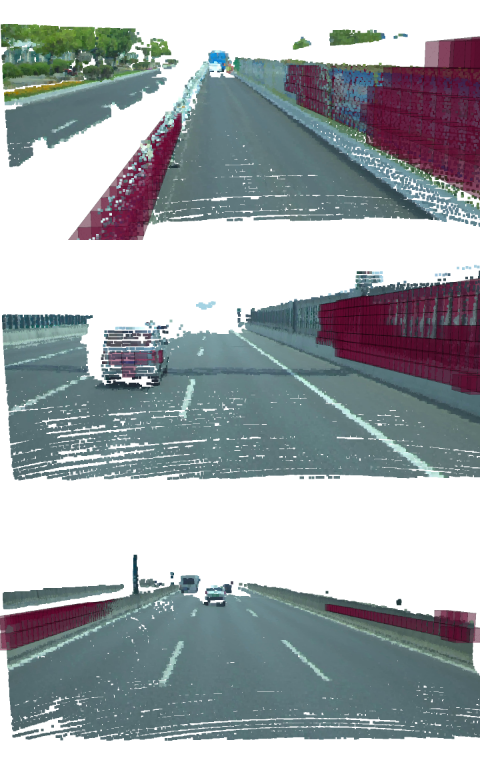}
\textbf{Ours}
\end{minipage}
\caption{Visual results of obstacle detection on the DrivingStereo dataset.}
\end{figure*}

\begin{table}[t!]
\vspace*{0.15cm}
    \caption{
    \textbf{Quantitative obstacle detection results on the DrivingStereo testing set.} The best result is bold and the second-best result is underlined.
    We use an RTX A5000 GPU for measuring the inference speed. 
    }
    \label{tab:occupancy}
    \centering
    \resizebox{\columnwidth}{!}{%
    \begin{tabular}{l|c|r r|r r|r|r|r}
    \hline
    \multirow{2}{*}{Method} & \multirow{2}{*}{Level} & \multicolumn{2}{c|}{CD (meters) $\downarrow$} & \multicolumn{2}{c|}{IoU (\%) $\uparrow$} & \multirow{2}{*}{MACs $\downarrow$} & \multirow{2}{*}{Params $\downarrow$} & \multirow{2}{*}{FPS $\uparrow$} \\
    & & 15.0m & 30.0m & 15.0m & 30.0m & & \\
    \hline
    2D-MSNet~\cite{shamsafar_mobilestereonet_2022} & 4 & 9.72 & 21.01 & 18.14 & 9.38 & 91.74G & \underline{2.23M} & 6.7 \\
    3D-MSNet~\cite{shamsafar_mobilestereonet_2022} & 4 & 5.96 & 13.14 & 21.11 & 11.96 & 414.59G & \textbf{1.77M} & 3.6 \\
    ACVNet~\cite{xu_attention_2022} & 4 & 8.43 & 18.82 & 18.32 & 13.59 & 801.33G & 7.11M & 4.4 \\
    Lac-GwcNet~\cite{liu_local_2022} & 4 & 7.19 & \underline{12.90} & 29.84 & 17.25 & 777.19G & 9.37M & 4.8 \\
    CFNet~\cite{shen_cfnet_2021} & 4 & 9.79 & 18.65 & 21.21 & 10.35 & 456.14G & 22.24M & 4.9 \\
    \hline
    \multirow{4}{2.5cm}{StereoVoxelNet~\cite{li2023stereovoxelnet}} & 1 & 4.54 & 5.12 & 80.07 & 71.94 & \multirow{4}{*}{\underline{16.03G}} & \multirow{4}{*}{4.58M} & \multirow{4}{*}{\textbf{21.8}} \\
                                    & 2 & 3.07 & 4.34 & 70.08 & 56.39 & & \\
                                    & 3 & 2.75 & 6.87 & 51.80 & 37.14 & & \\
                                    & 4 & 3.15 & 17.20 & 38.27 & 21.92 & & \\
    \hline
    \multirow{4}{2.5cm}{StereoVoxelNet \\ w/ EfficientNet} & 1 & 4.36 & 5.17 & 80.26 & 71.93 & \multirow{4}{*}{\textbf{14.83G}} & \multirow{4}{*}{10.46M} & \multirow{4}{*}{20.2} \\
                                    & 2 & 2.81 & 4.28 & 70.55 & 56.32 & & \\
                                    & 3 & 2.49 & 5.95 & 52.59 & 37.83 & & \\
                                    & 4 & \underline{2.90} & 13.56 & \underline{38.69} & \underline{23.04} & & \\
    \hline
    \multirow{4}{2.5cm}{ODTFormer (\textbf{Ours})} & 1 & \textbf{3.09} & \textbf{4.13} & \textbf{85.25} & \textbf{76.75} & \multirow{4}{*}{25.05G} & \multirow{4}{*}{6.14M} & \multirow{4}{*}{\underline{21.7}} \\
                                    & 2 & \textbf{1.92} & \textbf{3.22} & \textbf{76.10} & \textbf{62.69} & & \\
                                    & 3 & \textbf{1.64} & \textbf{3.82} & \textbf{57.94} & \textbf{43.79} & & \\
                                    & 4 & \textbf{1.87} & \textbf{8.72} & \textbf{41.62} & \textbf{26.62} & & \\
    \hline
    \end{tabular}
    }
\end{table}

\subsection{Obstacle Detection}
\label{sec: ob-detection}

We optimize our voxel occupancy prediction module using the DrivingStereo dataset~\cite{yang2019drivingstereo}, which contains over $180,000$ stereo pairs for the driving scenario with a resolution of $400 \times 880$. 
During preprocessing, we augment the training images using color jittering.

We set our ROI (defined in Sec.~\ref{sec: occ_decoding}) with respect to the left camera as $30m$ ahead, $8m$ to the left, $10m$ to the right, and $3m$ to the top and bottom. 
We set the voxel sizes to be $3m$, $1.5m$, $0.75m$, and $0.375m$ for each level, and the finest level will have a voxel grid of size $n_x \times n_y \times n_z = 48 \times 16 \times 80$.
We precompute and store the precise voxel occupancy grid ground truth by cropping our ROI and removing all voxels that are $1.5m$ under the viewpoint on the $y$-axis (ground plane).

The network is trained for 20 epochs with a batch size of 16 using AdamW~\cite{loshchilov_decoupled_2018} optimizer. 
We set the initial learning rate as $0.0001$ and decrease it each epoch using the Cosine Annealing learning rate scheduler with a minimum learning rate of $1 \times 10^{-8}$.

\textbf{Evaluation metrics.} 
We choose IoU and Chamfer Distance (CD) as our evaluation metrics.
IoU measures the intersection between the ground truth and the predicted voxel occupancy grids.
CD measures the distance between the point clouds (by presenting the centroids of occupied voxels as point clouds).
We report the output accuracies with the $15.0m$ range (half of the ROI depth) and the $30.0m$ range (entire ROI depth) separately, which represents easier and harder-to-identify obstacles. 

Besides the accuracy metrics, we also consider computational cost matrics: multiply-accumulate operations (MACs), the number of model parameters (Params), and frame-per-second (FPS). 

\textbf{Baseline methods.} 
We compare obstacle detection performance against two types of seminal works: stereo-based depth estimation and voxel occupancy prediction.
For depth estimation models, following the prior works~\cite{eppenberger_leveraging_2020, li2023stereovoxelnet}, we first convert the estimated depth map into a point cloud and subsequently voxelize it into a voxel occupancy grid.
We include MobileStereoNet (MSNet, both 2D and 3D version)~\cite{shamsafar_mobilestereonet_2022}, ACVNet~\cite{xu_attention_2022}, Lac-GwcNet~\cite{liu_local_2022} and CFNet~\cite{shen_cfnet_2021}.
For the occupancy prediction approach, we compare against StereoVoxelNet~\cite{li2023stereovoxelnet}.
To rule out the effect of feature extraction, we also ablate the original feature extraction backbone of StereoVoxelNet with EfficientNet (same as ours).

As shown in \cref{tab:occupancy}, ODTFormer significantly outperforms all other approaches, particularly depth-based methods~\cite{shamsafar_mobilestereonet_2022, xu_attention_2022, liu_local_2022, shen_cfnet_2021}, in terms of both IoU and CD.
Although ODTFormer exhibits higher MACs compared to StereoVoxelNet, we achieve similar FPS. 
This discrepancy arises because we do not construct cost volumes iteratively based on stereo matching, enabling better parallelism.

\subsection{Obstacle Tracking}
\label{sec: ob-tracking}

\begin{figure*}[t!]
\vspace*{0.15cm}
\noindent
\centering
\begin{tabularx}{\textwidth}{c *{5}{>{\centering\arraybackslash}X}}
    
    \rotatebox[origin=c]{90}{Input} &
    \raisebox{-0.5\height}{\includegraphics[width=.95\textwidth]{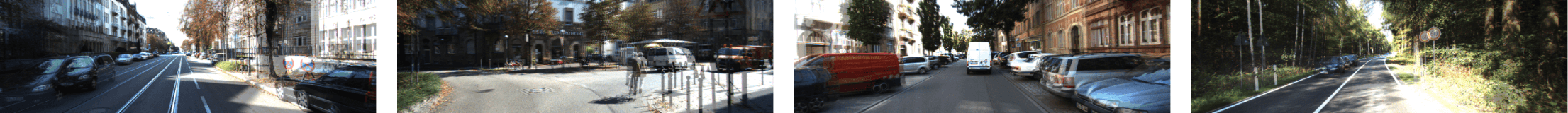}} \\

    \rotatebox[origin=c]{90}{GT} &
    \raisebox{-0.5\height}{\includegraphics[width=.95\textwidth]{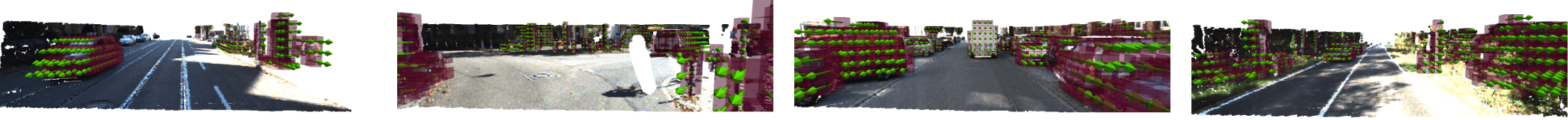}} \\
    
    \rotatebox[origin=c]{90}{PWOC} &
    \raisebox{-0.5\height}{\includegraphics[width=.95\textwidth]{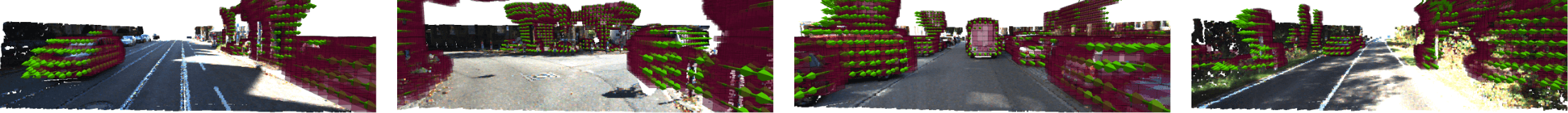}} \\

    \rotatebox[origin=c]{90}{RAFT} &
    \raisebox{-0.5\height}{\includegraphics[width=.95\textwidth]{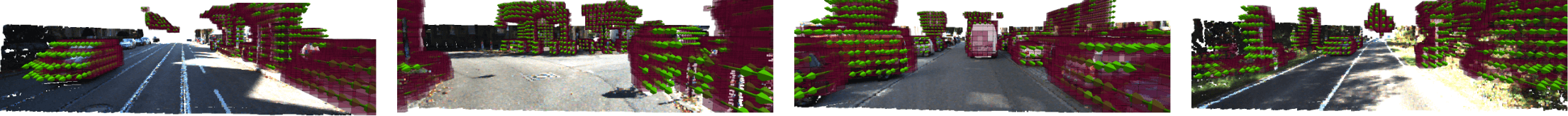}} \\

    \rotatebox[origin=c]{90}{Ours} &
    \raisebox{-0.5\height}{\includegraphics[width=.95\textwidth]{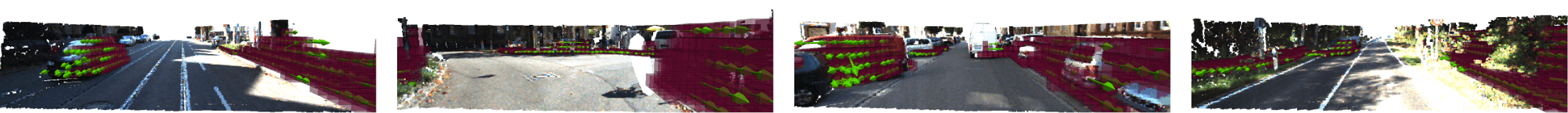}} \\
\end{tabularx}
\caption{
\textbf{Visual results of obstacle tracking on the KITTI 2015 dataset.}
The first row shows the stacked images from consecutive frames.
The obstacle detection results are shown as red cubes, and the tracking results are marked as green arrows.
Longer arrows indicate large motion magnitude.
}
\end{figure*}

\begin{table}[t!]
    \caption{
    \textbf{Quantitative obstacle tracking results on the KITTI 2015 scene flow dataset.}
    RAFT-3D MACs are evaluated including MobileStereoNet.
    The best result is bold, and the second-best result is underlined.
    }
    \label{tab:flow}
    \centering
    \resizebox{\columnwidth}{!}{%
    \begin{tabular}{l|r|r|r}
    \hline
    Method  & EPE (meters) $\downarrow$ & Foreground EPE (meters) $\downarrow$ & MACs $\downarrow$ \\
    \hline
    PWOC-3D~\cite{saxena2019pwoc} & 0.043 & 1.23 & N/A \\
    SENSE~\cite{jiang2019sense} & 0.048 & 1.18 & \underline{284.73G} \\
    RAFT-3D (2D-MSNet)~\cite{teed_raft-3d_2021} & \underline{0.025} & \underline{0.91} & 649.46G \\
    RAFT-3D (3D-MSNet)~\cite{teed_raft-3d_2021} & 0.028 & \textbf{0.87} & 1295.16G \\
    \hline
    ODTFormer (\textbf{Ours}) & \textbf{0.021} & 1.08 & \textbf{64.47G} \\
    \hline
    \end{tabular}
    }
\end{table}

We finetune ODTFormer for voxel tracking using the synthetic SceneFlow Driving dataset~\cite{mayer2016large} and the real-world KITTI 2015 dataset~\cite{menze2015object}, which contains over $4,000$ stereo image sequences with the resolution of $540 \times 960$ and 200 stereo images with the resolution of $370 \times 1224$, respectively.
We resize the images and camera parameters to have the same resolution $400 \times 880$ as in DrivingStereo, then apply the same data augmentation as obstacle detection and voxelize the scene flow by averaging the scene flow vectors of the point cloud within occupied voxels.
We set the FPS and velocity thresholds for the matching boundary as $f = 26$ FPS and $v=33.3 m/s$ (or 120 $km/h$).
The volumetric region of each voxel during tracking is set to be the surrounding $9 \times 3 \times 9$ voxels along $x$, $y$, $z$ axes.

We randomly split the KITTI training dataset into a custom training set and a validation set using an 8:2 ratio. 
We first train the network for 20 epochs with a batch size of 8 on the SceneFlow Driving dataset, then finetune on the KITTI dataset for 50 epochs.

\textbf{Evaluation metrics.} 
We employ EPE and foreground EPE to assess our model's performance in voxel flow estimation. 
EPE quantifies the Euclidean distance between the ground-truth flow vector and the estimated flow vector.
Foreground EPE is the EPE loss measured upon those voxels classified as occupied within the ground truth so that the measurement considers both occupancy detection and flow estimation accuracies.

\textbf{Baseline methods.} 
We compare against stereo-based scene flow estimation methods: SENSE~\cite{jiang2019sense}, PWOC-3D~\cite{saxena2019pwoc}, and RAFT-3D~\cite{teed_raft-3d_2021} with MobileStereoNet (MSNet)~\cite{shamsafar_mobilestereonet_2022} for depth estimation.
We use their model weights trained on the \emph{entire} KITTI 2015 training set, including our custom validation set, to evaluate all the methods. 
We first estimate the scene flow given stereo image pairs from consecutive frames, then voxelize it using the same process as we generate the ground truth (\cref{sec: ob-tracking}).

\textbf{Quantitative results.}
The quantitative results are evaluated using our custom validation set within the KITTI 2015 training data since our model cannot be directly evaluated through the KITTI test submission.
The results are summarized in Table \ref{tab:flow}.
To filter outliers in dense scene flow estimation methods, we clipped the results from compared methods into the same bounded tracking boundary of [9, 3, 9].
Although the validation set we use is included in the baseline methods' training data, with significantly fewer MACs than all dense scene flow methods, ODTFormer still outperforms baselines on both metrics except RAFT-3D.
It validates the effectiveness of our proposed approach to track obstacles.
Notably, we achieve comparable performance to RAFT-3D using only ten-fold or twenty-fold less MAC operations.
In terms of inference speed, the tracing module adds negligible burden to the detection module thanks to our design.
The entire model runs at 20fps for the KITTI resolution ($370 \times 1224$) on an RTX A5000 GPU without careful postprocessing such as quantization, pruning, etc.

\begin{table}[t!]
    \caption{Ablation study for obstacle detection.}
    \label{tab:ablation_occ}
    \centering
    \resizebox{\columnwidth}{!}{%
    \begin{tabular}{l|c|r r|r r|r|r}
    \hline
    \multirow{2}{*}{Method} & \multirow{2}{*}{Level} & \multicolumn{2}{c|}{CD (meters) $\downarrow$} & \multicolumn{2}{c|}{IoU (\%) $\uparrow$} & \multirow{2}{*}{MACs $\downarrow$} & \multirow{2}{*}{Params $\downarrow$} \\
    & & 15.0m & 30.0m & 15.0m & 30.0m & & \\
    \hline
    \multirow{4}{3cm}{Ours w/o geometric constraints} & 1 & 6.17 & 9.92 & 77.77 & 64.01 & \multirow{4}{*}{\textbf{23.79G}} & \multirow{4}{*}{7.52M} \\
                                    & 2 & 4.69 & 9.34 & 66.94 & 47.36 & & \\
                                    & 3 & 4.56 & 12.97 & 48.58 & 29.40 & & \\
                                    & 4 & 4.98 & 23.65 & 34.07 & 17.10 & & \\
    \hline
    \multirow{4}{3cm}{Ours w/o 3D volumetric sampling} & 1 & 3.73 & 5.49 & 82.53 & 72.92 & \multirow{4}{*}{24.65G} & \multirow{4}{*}{6.22M} \\
                                    & 2 & 2.51 & 4.53 & 72.66 & 57.72 & & \\
                                    & 3 & 2.23 & 5.74 & 54.18 & 38.45 & & \\
                                    & 4 & 2.42 & \textbf{12.03} & 38.30 & 22.57 & & \\
    \hline
    \multirow{4}{2.5cm}{Ours} & 1 & \textbf{3.09} & \textbf{4.13} & \textbf{85.25} & \textbf{76.75} & \multirow{4}{*}{25.05G} & \multirow{4}{*}{\textbf{6.14M}} \\
                                    & 2 & \textbf{1.92} & \textbf{3.22} & \textbf{76.10} & \textbf{62.69} & \\
                                    & 3 & \textbf{1.64} & \textbf{3.82} & \textbf{57.94} & \textbf{43.79} & \\
                                    & 4 & \textbf{1.87} & \textbf{8.72} & \textbf{41.62} & \textbf{26.62} & \\
    \hline
    \end{tabular}
    }
\end{table}

\subsection{Ablation Studies}
\label{sec: ablation}

\textbf{Obstacle prediction.} 
To study the effectiveness of our design choices, we conduct ablation studies and report the results in Table~\ref{tab:ablation_occ} using the DrivingStereo dataset.

We first replace the feature concatenations in Eq.(\ref{eq:matching_cost}) and (\ref{eq:matching_cost_distill}) with cross-view averaging as in existing work~\cite{li_bevformer_2022, li_voxformer_2023, wei_surroundocc_2023, huang_tri-perspective_2023}.
In this way, it discards the geometric constraints between two stereo images.
As we can see at the top of Table~\ref{tab:ablation_occ}, it leads to significantly worse detection accuracy.

We also study the effectiveness of the matching cost refinement and simply use $\rmC_B$ as the final cost volume.
As shown in the middle of Table~\ref{tab:ablation_occ}, it shows worse results than our full model with fewer model parameters and a slight decrease in computation burden.
Given the fast inference speed of our model, we include it for the best obstacle detection accuracy.

\textbf{Voxel tracking.}
The ablation study for our voxel tracker design is reported in Table \ref{tab:ablation_track} using KITTI 2015. 
We see that without the bounded tracking design, the tracker directly results in an out-of-memory error when evaluating with an RTX A5000 GPU.
Whether to match occupied voxels in the second frame is an open design choice. 
It leads to a minor performance degradation in the foreground area,
partially because the occupancy detection results may not always be correct.
We therefore opt to track all voxels for the best tracking accuracy.

\begin{table}[t!]
    \caption{\textbf{Ablation study for obstacle tracking.} \texttt{OOM} indicates an out-of-memory error.
    }
    \label{tab:ablation_track}
    \centering
    \resizebox{\columnwidth}{!}{%
    \begin{tabular}{c|c|c|c}
    \hline
    \multirow{2}{*}{Bounded} & Match Occupied Voxels Only & EPE$\downarrow$ & Foreground EPE$\downarrow$ \\
     & in the Second Frame & (meters) & meters \\
    \hline
    \xmark & \xmark & \texttt{OOM} & \texttt{OOM} \\
    \cmark & \cmark & \textbf{0.021} & 1.092 \\
    \cmark & \xmark & \textbf{0.021} & \textbf{1.087}\\
    \hline
    \end{tabular}
    }
\end{table}

\section{Conclusion}
This paper contributed ODTFormer, a Transformer-based model to address obstacle detection and tracking problems.
We detect the obstacles in the form of a voxel occupancy grid using a deformable attention mechanism and track them by matching voxels across two consecutive frames.
We confirmed the effectiveness of ODTFormer by providing extensive quantitative and qualitative results on DrivingStereo and KITTI datasets.
Our results showed that ODTFormer achieves state-of-the-art performance on obstacle detection tasks.
For the obstacle tracking task, we showed that ODTFormer can achieve accuracy comparable to state-of-the-art methods.
The entire model is efficient and runs fast at 20fps for the KITTI resolution ($370 \times 1224$) on an RTX A5000 GPU without careful postprocessing (\eg, quantization, pruning).
Being able to detect and track obstacles at the same time will empower a set of downstream tasks, for instance, ensuring safe navigation when people move around a robot.

\footnotesize{
\bibliographystyle{IEEEtranN}
\bibliography{hongyu_zotero, custom}
}
\end{document}